\documentclass[preprint,12pt]{elsarticle}

\usepackage[utf8]{inputenc} 
\usepackage{graphicx}
\usepackage{amsmath, amssymb} 
\usepackage{hyperref}
\usepackage{orcidlink}

\begin{document}

\begin{frontmatter}

\title{Acoustic Index: A Novel AI-Driven Parameter for Cardiac Disease Risk Stratification Using Echocardiography}

\author[1]{Beka Begiashvili\,\orcidlink{0000-0002-9993-0915}}
\ead{beka.begiashvili@corsonic.com}
\author[1]{Carlos J. Fernandez-Candel\,\orcidlink{0000-0002-3835-9428}}
\author[1,2]{Matías Pérez Paredes}

\address[1]{Corsonic, Murcia, Spain}
\address[2]{Hospital Morales Meseguer, Murcia, Spain}

\begin{abstract}
Traditional echocardiographic parameters such as ejection fraction (EF) and global longitudinal strain (GLS) have limitations in the early detection of cardiac dysfunction. EF often remains normal despite underlying pathology, and GLS is influenced by load conditions and vendor variability. There is a growing need for reproducible, interpretable, and operator-independent parameters that capture subtle and global cardiac functional alterations. 

We introduce the Acoustic Index, a novel AI-derived echocardiographic parameter designed to quantify cardiac dysfunction from standard ultrasound views. The model combines Extended Dynamic Mode Decomposition (EDMD) based on Koopman operator theory with a hybrid neural network that incorporates clinical metadata. Spatiotemporal dynamics are extracted from echocardiographic sequences to identify coherent motion patterns. These are weighted via attention mechanisms and fused with clinical data using manifold learning, resulting in a continuous score from 0 (low risk) to 1 (high risk). 

In a prospective cohort of 736 patients, encompassing various cardiac pathologies and normal controls, the Acoustic Index achieved an area under the curve (AUC) of 0.89 in an independent test set. Cross-validation across five folds confirmed the robustness of the model, showing that both sensitivity and specificity exceeded 0.8 when evaluated on independent data, particularly at a threshold of 0.45 for the Acoustic Index. Threshold-based analysis demonstrated stable trade-offs between sensitivity and specificity, with optimal discrimination near this threshold.

The Acoustic Index represents a physics-informed, interpretable AI biomarker for cardiac function. It shows promise as a scalable, vendor-independent tool for early detection, triage, and longitudinal monitoring. Future directions include external validation, longitudinal studies, and adaptation to disease-specific classifiers. 
\end{abstract}


\end{frontmatter}

\section{Introduction}

Cardiovascular disease (CVD) remains the foremost cause of mortality worldwide, placing enormous strain on healthcare systems and underscoring the need for efficient and accurate diagnostic tools. Global demographic shifts, notably the rising proportion of older adults, are further amplifying the incidence of heart failure, valvular disorders, and ischemic heart disease, thereby driving up the demand for cardiac imaging. This situation calls for methods capable of detecting early or subclinical cardiac dysfunction, particularly in at-risk patients such as those with multiple comorbidities or undergoing therapies known to affect cardiac function. Echocardiography has emerged as the frontline modality for cardiac assessment due to its noninvasive nature, wide availability, and ability to visualize key structural and functional parameters of the heart in real time. Despite these advantages, contemporary literature stresses that the escalating prevalence of CVD, coupled with an increasingly elderly population, requires more robust and operator-independent echocardiographic measures to ensure early detection and timely intervention \cite{Brady2023, Zhao2024, Strait2012}.\\

While traditional indices like ejection fraction (EF) and global longitudinal strain (GLS) have guided diagnostic and treatment decisions for decades, both metrics have intrinsic shortcomings that can hinder their clinical utility in detecting subtle or evolving cardiac abnormalities. EF often remains within normal limits until significant myocardial damage has occurred, rendering it less sensitive to early pathological changes. Meanwhile, GLS is more sensitive to subclinical dysfunction but is notably influenced by variations in loading conditions and by the proprietary algorithms of different ultrasound vendors, which can lead to inconsistent values between devices. Both EF and strain assessments are further limited by their reliance on high-quality image acquisition, which is subject to inter-operator variability and can be hampered by patient-specific factors such as obesity or poor acoustic windows. Collectively, these issues reduce reproducibility, impose longer scanning times, and leave room for missed diagnoses in the early stages of disease—prompting calls for innovative echocardiographic solutions that minimize user dependence and capture a broader spectrum of functional indicators \cite{Brady2023, Zhao2024, Strait2012, Manar2019}.

\subsection{Artificial Intelligence-Based Echocardiography and Dynamical Systems Theory}
Artificial intelligence, particularly deep learning, has significantly reshaped echocardiography \cite{Alsharqi2018} by enhancing measurement reproducibility and automating tasks like chamber segmentation \cite{Assadi2024, AlKindi2021, Arafati2020}, view classification \cite{Park2007, Balaji2015, Gao2017}, and even estimation of left ventricular ejection fraction (LVEF) \cite{Reynaud2021, SanchezMartinez2018}. Advanced video-based neural networks now analyze entire cardiac cycles \cite{Charitha2023}, thereby reducing operator dependence and improving the detection of early-stage disease. Other machine learning models integrate multiple echo-derived metrics, such as chamber dimensions, Doppler velocities, and strain parameters, to predict patient outcomes with greater accuracy than EF alone. However, a lack of interpretability often hinders these methods; most deep neural networks behave as ``black boxes,” making it difficult for clinicians to validate how input features contribute to a given risk classification or diagnosis. Recent research into explainable artificial intelligence aims to address these concerns by providing more transparent decision pathways while retaining the models’ predictive power \cite{Ouyang2020, Akkus2021}.\\

To surmount the limits of purely data-driven approaches, researchers are increasingly turning to dynamical systems theory, most notably Koopman operator-based techniques, to capture the underlying temporal structure of beating-heart data. By viewing an echocardiogram as a nonlinear system evolving over time, Koopman decomposition methods can separate distinct oscillatory modes (e.g., cardiac contraction-relaxation frequencies, respiratory signals) and thus highlight subtle, physiologically meaningful patterns. Early applications in animal models have demonstrated the feasibility of discerning pathologic states like myocardial infarction or hypertrophic cardiomyopathy through the identification of abnormal dynamic “fingerprints.” These dynamical approaches offer a critical advantage over static or frame-based artificial intelligence solutions, as they explicitly factor in how regional motion evolves, opening the door to improved classification of subclinical dysfunction. When integrated with machine learning pipelines, Koopman-derived features enable low-dimensional yet interpretable representations of complex echo data, paving the way for robust, physics-informed models that promise greater reliability and transparency in clinical echocardiography \cite{Groun2022, Groun2023, Zhang2019}.


\subsection{The Proposed Innovation: The Acoustic Index}
Building on current advances data-driven echocardiography and the insights from dynamical systems theory, the Acoustic Index (AI) aims to unify traditional echocardiographic measures (e.g., ejection fraction, strain) with computationally derived spatiotemporal features into one interpretable metric. Rather than focusing on isolated frames or single-parameter outputs, the Acoustic Index leverages Extended Dynamic Mode Decomposition\cite{Williams2015} (EDMD) to capture the temporal evolution of the beating heart. This approach isolates distinct oscillatory components, such as the fundamental contraction–relaxation cycle or subtle respiratory-induced shifts, while filtering out noise and artifacts. A deep neural network subsequently integrates these modes with conventional parameters and additional clinical inputs, 
producing a single score indicative of overall cardiac function. By prioritizing the detection of subtle motion patterns, the Acoustic Index can flag subclinical dysfunction even when EF remains normal. Additionally, its hybrid architecture (i.e., Koopman-based feature extraction plus machine learning) helps reduce operator dependence, enhance reproducibility, and maintain transparency: interpretable output ``modes” allow to trace back which dynamic patterns most strongly influenced the score.\\

Clinically, the Acoustic Index could address the need for more efficient triaging of patients in busy echo labs or resource-limited settings. Since it consolidates multiple echo-derived variables into one numeric value, physicians can spot deteriorating function at a glance and prioritize further imaging or intervention accordingly. The Acoustic Index also holds promise for monitoring therapy response, guiding medication adjustments, or scheduling follow-up based on a trajectory of functional change over time. Moreover, unlike black-box Artificial Intelligence systems, the Acoustic Index provides interpretable elements at each stage, from raw echocardiography loops to the final composite score, thereby building clinician confidence and fostering adoption. As research accumulates to validate its prognostic value across diverse patient populations, this unified spatiotemporal clinical parameter may redefine how echocardiography is harnessed for early detection, risk stratification, and cardiac care.

\section{Methodology \label{sec:methodology}}

This section describes the comprehensive framework underpinning the Acoustic Index, a novel AI-driven clinical parameter for cardiac risk stratification. 
We first detail the data acquisition and preprocessing pipeline. 
Next, we introduce the theoretical foundations, integrating data-driven Koopman operator theory, specifically Extended Dynamic Mode Decomposition adoption (EDMD), with deep learning to model spatiotemporal cardiac dynamics. The hybrid architecture synthesizes these dynamical modes with traditional clinical parameters (e.g., ejection fraction, age) through attention mechanisms and manifold learning. 
Finally, we derive the Acoustic Index mathematically, describing its components, clinical interpretability, and proprietary safeguards. 
Validation protocols, including train-test splits and benchmarking against established metrics, are described to ensure reproducibility while protecting intellectual property. 
This methodology bridges dynamical systems theory, computational imaging, and clinical cardiology, establishing a rigorous yet translatable framework for AI-enhanced echocardiography.




\subsection{Theoretical Framework}  

\subsubsection{Koopman Operator for Cardiac Dynamics}  
The human heart exhibits highly nonlinear and time-dependent motion, governed by complex interactions between myocardial contraction, chamber geometry, and hemodynamics. Traditional models often simplify this behavior through local linearization, limiting their ability to detect early-stage or spatially diffuse dysfunctions.\\

To overcome these limitations, we employ the \textit{Koopman operator} framework, which offers a linear but infinite-dimensional representation of nonlinear dynamical systems. Instead of directly modeling the evolution of the system state $x(t)$, the Koopman operator $\mathcal{K}$ advances a set of observable functions $g(x)$, such that:
\[
\mathcal{K}g(x(t)) = g(x(t + \Delta t)),
\]
where $g$ may represent any measurable feature of the cardiac system, including displacement fields, flow velocities, or strain patterns. By lifting the dynamics into the space of observables, the Koopman operator enables analysis of complex spatiotemporal behavior using linear techniques, crucially without requiring local approximations.\\

This operator-theoretic approach allows us to identify recurrent temporal structures and spatially localized instabilities from echocardiographic sequences. In the context of cardiac imaging, it facilitates the detection of subtle dynamic anomalies, such as early diastolic dysfunction or regional dyssynchrony, that may not be captured by conventional frame-based metrics like ejection fraction.

\subsubsection{Modal Decomposition via Extended Dynamic Mode Decomposition (EDMD)}  
To approximate the Koopman operator in practice, we use \textit{Extended Dynamic Mode Decomposition} (EDMD), a data-driven method that extracts coherent spatiotemporal patterns from sequential ultrasound frames. This approach, originally formalized by Williams et al.~\cite{Williams2015}, enables interpretable decomposition of cardiac motion into dynamic modes.

Given a sequence of frames $\{x_t\}_{t=1}^{T}$, each frame is lifted into a higher-dimensional feature space via a dictionary of nonlinear basis functions $\Psi(x) = [\psi_1(x), \ldots, \psi_m(x)]^T$. From this, we construct the matrices:
\[
\mathbf{\Psi} = [\Psi(x_1), \ldots, \Psi(x_{T-1})], \quad \mathbf{\Psi}' = [\Psi(x_2), \ldots, \Psi(x_T)],
\]
and compute the finite-dimensional Koopman matrix:
\[
\mathbf{K} = \mathbf{\Psi}' \mathbf{\Psi}^\dagger,
\]
where $\mathbf{\Psi}^\dagger$ is the Moore--Penrose pseudoinverse. The eigendecomposition of $\mathbf{K}$ yields:

\begin{itemize}
    \item \textbf{Eigenvalues} $\lambda_i$, representing the temporal behavior of dynamic patterns (growth, decay, oscillation).
    \item \textbf{Eigenvectors} $v_i$, defining Koopman modes in the lifted observable space.
    \item \textbf{Eigenfunctions} $\varphi_i(x) \approx \Psi(x)^T v_i$, providing scalar-valued functions that evolve linearly in time: $\varphi_i(x_{t+1}) \approx \lambda_i \varphi_i(x_t)$.
\end{itemize}

These components form the spectral backbone of our system. The \textit{eigenvalues} $\lambda_i$ are used to assess the \textit{stability} of motion patterns; modes with $\text{Re}(\lambda_i) > 0$ are treated as indicators of pathologic growth or instability (e.g., arrhythmogenic motion). The \textit{eigenfunctions} $\varphi_i(x)$ are used to localize these patterns anatomically, with further clinical relevance assessed through metrics such as the Sobolev norm $\|\varphi_i\|_{H^1}$ for spatial irregularity and $\|\partial_t \varphi_i\|^2$ for temporal fluctuations.

We deliberately adopt EDMD over kernelized or deep Koopman variants to ensure \textit{clinical interpretability}, \textit{transparency}, and \textit{spatial resolution}, all of which are essential for translational applicability in echocardiography.

\subsubsection{Neural Network Architecture}
A hybrid model refines EDMD-derived modes while integrating clinical parameters:
\begin{enumerate}
    \item Encoder: EDMD preprocesses DICOM sequences into \(k\) spatiotemporal modes. A proprietary spectral filter removes noise-dominated modes (e.g., speckle artifacts).
    \item Attentional Gating: An attention layer assigns saliency weights \(w_i\) to each mode:  
    \[
    w_i = \text{softmax}\left( \mathbf{Q}^T \sigma(\mathbf{V}[\varphi_i; \lambda_i]) \right),
    \]
    where \(\mathbf{Q}, \mathbf{V}\) are learned matrices, and \(\sigma\) is a Swish activation. This prioritizes modes with high instability (\(\text{Re}(\lambda_i) > 0\)) or anatomical specificity.
    \item Manifold Fusion: Clinical parameters \(\mathbf{p} = [EF, D, \text{age}, \text{sex}]\) are embedded into a latent manifold \(\mathcal{M}\) using a Siamese network. Geodesic distances on \(\mathcal{M}\) model nonlinear interactions between clinical and dynamical features (e.g., age-modulated mode persistence). 
\end{enumerate}

\subsection{Methodological Integration}
This framework integrates three pillars of analysis to transform echocardiography into a predictive clinical tool. 
First, Koopman operator theory and Extended Dynamic Mode Decomposition (EDMD) extract interpretable spatio-temporal modes from raw ultrasound sequences, capturing transient dynamics such as dyssynchrony or diastolic instability. 
These modes are refined through a hybrid neural network that prioritizes clinically relevant patterns using attention mechanisms, ensuring computational efficiency and alignment with cardiologists’ diagnostic intuition. 
Concurrently, traditional clinical parameters (ejection fraction, age, ventricular dimensions) are nonlinearly embedded into a latent space, where their interactions with dynamical modes are modeled via manifold learning. 
This fusion enables the system to contextualize instability, for example, linking age-related myocardial stiffness to specific aberrant motion patterns. 
The pipeline’s modular design ensures adaptability across populations while preserving interpretability: clinicians can trace risk contributions to unstable modes or clinical factors. Proprietary safeguards, including mode filtering and manifold parameterization, protect intellectual property without compromising reproducibility. 
By unifying dynamical systems theory, deep learning, and clinical expertise, the methodology advances echocardiography from qualitative imaging to quantitative, AI-driven risk stratification.

\subsection{Acoustic Index Derivation \label{sec:acoustic_index_derivation}}
The Acoustic Index (\(AI\)) is formulated as a probabilistic synthesis of dynamical instability, clinical risk, and their nonlinear interactions, derived through a hierarchical fusion of Koopman operator theory and deep learning. Let \(\{\varphi_i, \lambda_i\}_{i=1}^k\) denote the Koopman modes and eigenvalues extracted via EDMD, and let \(\mathbf{p} \in \mathbb{R}^d\) represent normalized clinical parameters (e.g., ejection fraction, age). The Acoustic Index is defined as:  

\[
AI = \Phi\left(\, \underbrace{\sum_{i=1}^k w_i \cdot \mathcal{D}(\lambda_i, \varphi_i)}_{\text{Dynamical Risk}} \,+\, \underbrace{\mathcal{C}(\mathbf{p})}_{\text{Clinical Risk}} \,+\, \underbrace{\mathcal{I}(\varphi_i, \mathbf{p})}_{\text{Cross-Domain Interaction}} \, \right),
\]  

where:
\begin{itemize}
    \item \(\Phi(z) = \frac{1}{1 + e^{-\eta(z - z_0)}}\) is a calibrated sigmoid with curvature \(\eta > 0\) and threshold \(z_0\), ensuring \(AI \in [0,1]\).
    \item \(\mathcal{D}(\lambda_i, \varphi_i)\) quantifies the dynamical risk of mode \(i\):
    \[
    \mathcal{D}(\lambda_i, \varphi_i) = \text{Re}(\lambda_i) \cdot \underbrace{\|\varphi_i\|_{\mathcal{H}^1}}_{\substack{\text{Spatial} \\ \text{Inhomogeneity}}} + \underbrace{\mathcal{P}(\varphi_i)}_{\substack{\text{Temporal} \\ \text{Irregularity}}},
    \]
    \begin{itemize}
        \item \(\|\varphi_i\|_{\mathcal{H}^1}\): Sobolev norm penalizing non-smooth spatial patterns (e.g., scar boundaries).
        \item \(\mathcal{P}(\varphi_i) = \int_{0}^{T} \left|\frac{\partial \varphi_i}{\partial t}\right|^2 dt\): Integrated temporal gradient, capturing erratic motion (e.g., arrhythmic fluctuations).
    \end{itemize}
    \item \(\mathcal{C}(\mathbf{p})\) encodes clinical risk through nonlinear transformations of traditional parameters:  
    \[
    \mathcal{C}(\mathbf{p}) = \mathbf{\beta}^T \cdot \mathbf{p} + \mathbf{p}^T \mathbf{\Omega} \mathbf{p},
    \]  
    where \(\mathbf{\beta} \in \mathbb{R}^d\) and \(\mathbf{\Omega} \in \mathbb{R}^{d \times d}\) are learned coefficients encoding linear and quadratic risk interactions (e.g., \(\beta_{\text{EF}} < 0\) links low ejection fraction to risk; \(\Omega_{\text{age,EF}} > 0\) models accelerated risk in elderly patients with reduced EF).
    \item \(\mathcal{I}(\varphi_i, \mathbf{p})\) represents cross-domain interactions, computed as:  
    \[
    \mathcal{I}(\varphi_i, \mathbf{p}) = \sum_{i=1}^k \text{Re}(\lambda_i) \cdot \text{dist}_{\mathcal{M}}(\varphi_i, \mathbf{p}),
    \]  
    where \(\text{dist}_{\mathcal{M}}(\cdot)\) measures geodesic distances on a latent manifold \(\mathcal{M}\) that embeds both dynamical modes and clinical parameters. This term quantifies synergies (e.g., unstable apical modes \(\varphi_i\) coupled with advanced age amplify risk multiplicatively).\\
\end{itemize}

Clinical interpretability emerges from three interrelated components. 
First, \textbf{Dynamical Risk} leverages Koopman modes with \(\text{Re}(\lambda_i) > 0\) to signify pathological growth (e.g., dyssynchrony), while \(\|\varphi_i\|_{\mathcal{H}^1}\) localizes the anatomical region of dysfunction (e.g., septal vs. apical). 
Furthermore, high \(\mathcal{P}(\varphi_i)\) indicates elevated arrhythmic risk. 
Second, \textbf{Clinical Risk} captures compounding factors through quadratic terms \(\mathbf{p}^T \mathbf{\Omega} \mathbf{p}\), accommodating profiles such as older female patients with diastolic dysfunction. 
Finally, \textbf{Cross-Domain Interaction} utilizes geodesic distances on \(\mathcal{M}\) to illustrate how clinical variables, such as hypertension, can amplify certain pathological modes, thereby providing an integrated understanding of dynamical instability.\\

\textbf{Proprietary safeguards} ensure the protection of intellectual property at multiple levels. First, the exact parameterization of the manifold \(\mathcal{M}\), including its metric tensor and curvature, as well as details of the Siamese network architecture, remain undisclosed. Second, specific criteria for selecting the retained \(k\) modes, encompassing both spectral energy thresholds and clinical relevance, are considered proprietary to Corsonic. Finally, regularization strategies that penalize overfitting in \(\mathbf{\beta}\) and \(\mathbf{\Omega}\) are omitted from public disclosure.\\

Presented methodology integrates echocardiographic data from five standardized views: parasternal long-axis, parasternal short-axis (mid-papillary and aortic valve levels), apical 4-chamber, and apical 2-chamber into a unified computational framework. Each view undergoes parallel processing through Koopman operator-based modal decomposition, specifically using Extended Dynamic Mode Decomposition, to extract spatiotemporal modes that characterize cardiac motion dynamics. These modes encode stability or instability through their eigenvalues, with unstable modes (positive real components) flagged as potential markers of pathology. Spatial inhomogeneity and temporal irregularity of the modes are quantified using Sobolev norms and integrated temporal gradients, respectively, providing localized and time-sensitive risk metrics.\\ 

A hybrid neural architecture processes these dynamical features alongside clinical parameters such as ejection fraction, age, and ventricular dimensions. The network employs attention mechanisms to prioritize clinically relevant modes and views, followed by manifold learning to fuse dynamical and clinical data into a latent space. This space models nonlinear interactions, such as how age amplifies the risk contribution of specific motion anomalies. The fused risk score is mapped to a continuous probability via a calibrated sigmoid function, producing the Acoustic Index, which ranges from 0 (low risk) to 1 (high risk).\\

The Acoustic Index outputs a continuous score despite binary labels (healthy/disease) by learning probabilistic relationships between echocardiographic dynamics and clinical outcomes. During training, the model minimizes binary a loss function, treating the \(AI\) as the predicted probability of disease. The sigmoid function inherently maps the model’s raw output, a weighted combination of dynamical instability, clinical risk, and their interactions, to a [$0$,$1$] scale. This allows the \(AI\) to reflect gradations in risk even when trained on binary labels. For instance, subtle subclinical dysfunction in a ``healthy"-labeled patient may yield an intermediate \(AI\) (e.g., $0.4$), while severe pathology in a ``disease"-labeled patient approaches $0.9$. The model’s attention mechanisms and manifold fusion enable it to generalize beyond rigid labels, capturing early disease signatures or nuanced progression.\\

This formulation positions \textbf{Acoustic Index} as an \textbf{AI-based clinical parameter}, transcending conventional echocardiography by fusing dynamical systems theory, deep learning, and clinical expertise.

\subsection{Architecture of the Acoustic Index Computation Pipeline}

\begin{figure}[!h]
    \centering
    \includegraphics[width=\textwidth]{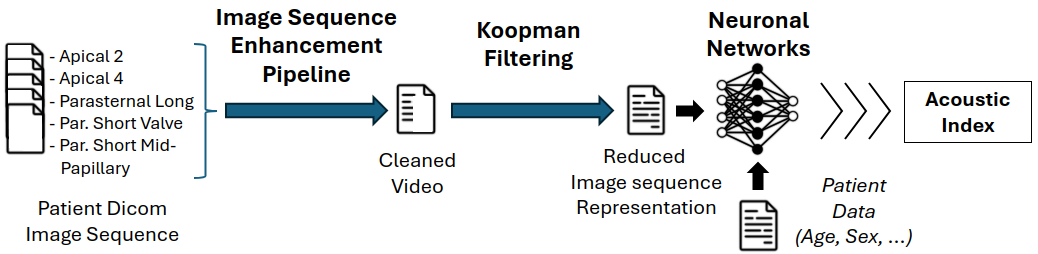}
    \caption{Architecture of the Acoustic Index Computation Pipeline.}
    \label{fig:architecture}
\end{figure}

The proposed pipeline is shown in Figure~\ref{fig:architecture}, which is organized into three major stages: Image Sequence Enhancement, the Koopman Filtering, and Neural Network Processing. 
The pipeline begins with the ingestion of raw DICOM ultrasound files corresponding to the five standardized cardiac views including Parasternal long-axis view (PLAX), Parasternal short-axis mid-papillary view (PSAX-MP), Parasternal short-axis aortic valve view (PSAX-AV), Apical 4-chamber view (A4C) and Apical 2-chamber view (A2C). These views collectively capture the anatomical (structural) and functional (dynamic) diversity of the heart. The input consists of ultrasound images' sequences stored in DICOM format, which include embed patient metadata such as age, sex, and other relevant clinical information, which are later used to complement the spatiotemporal-derived features in the final risk computation.

Each DICOM file undergoes a \textit{Image Sequence Enhancement Pipeline} designed to remove overlaid text, reduce frame noise, and optimize brightness and contrast, thereby improving the clarity of anatomical and motion-related features. In addition, all sequences are rescaled to a standardized spatial resolution and converted to a fixed aspect ratio, ensuring consistency across the rest of the pipeline. Therefore, this preprocessing step not only enhances visual quality but also harmonizes the data for subsequent stages, reducing inter-scan variability. 

The resulting image sequences are subjected to a temporal normalization process, designed to align and standardize the cardiac cycle phases (diastole and systole) across all input sequences. This step ensures that each sequence consistently represents a complete heartbeat cycle, beginning and ending at equivalent physiological states (end-diastole to end-diastole).

The enhanced sequences are then passed through the \textit{Koopman} filtering module (see previous section \ref{sec:acoustic_index_derivation} for theoretical details), which applies Koopman operator-based decomposition and proprietary spectral filtering. This step produces a reduced images' sequence representation that highlights both structural elements (e.g., myocardial borders, chamber geometry) and functional components (e.g., contraction-relaxation cycles). Additionally, this step reduces input dimensionality and model size, thereby improving computational efficiency and generalization.

The resulting representations are combined with patient metadata (e.g., age, sex, clinical metrics) and fed into a hybrid neural network that integrates both dynamic and static information. This fused input ultimately produces the Acoustic Index, a continuous value between 0 and 1 that reflects the patient's global cardiac risk. Although the general structure of the pipeline is described, the detailed architecture of the neural network is not publicly disclosed. This includes the number and type of layers, such as convolutional, normalization, or attention mechanisms, as well as activation functions and training parameters. These elements constitute proprietary intellectual property developed by Corsonic.

\subsection{Clinical Protocol and Dataset Annotation}  
This prospective observational study was approved by the Ethics Committee of Hospital Morales Meseguer (Murcia, Spain), following applicable data protection law and the Declaration of Helsinki. Patients $\ge 18$ years undergoing echocardiography between January–December 2024 were screened. Inclusion required acquisition of at least five standard views with Color Doppler: PLAX, PSAX-MP, PSAX-AV, A4C, A5C, and A2C, using Philips EPIQ or GE Vivid T8 systems at 30–60 fps.\\

Disease-negative cases were those reported as normal by the cardiologist performing the exam and confirmed by two independent reviewers. Disease-positive cases were initially identified by the acquiring cardiologist based on clinical and echo findings, and then validated by two observers. Assessment included 2D grayscale and Doppler evaluation of LV size, function, wall thickness; RV dimensions; atrial size; global/segmental systolic function; and valvular flow. 

\section{Results}

This section presents a structured evaluation of the Acoustic Index, combining statistical performance analysis with clinical interpretation. We begin with a summary of the dataset used for model development, including patient demographics, pathology types, and age distribution. We then describe the experimental methodology, which includes a five-fold cross-validation protocol followed by independent testing. The model's discriminative ability is assessed through ROC curves and AUC metrics across folds. To explore clinical applicability, we analyze how sensitivity and specificity vary with the decision threshold. Finally, we visualize the Acoustic Index at the individual patient level.

\subsection{Experimental Methodology}

To evaluate the learning capacity and generalization performance of the proposed Artificial Intelligence model, we conducted a two-stage experimental setup. First, the training set, comprising $624$ patients, was partitioned into five equally sized folds for internal cross-validation. In each of the $5$ iterations, $4$ folds ($80$\%) were used to train the hybrid neural network, while the remaining fold ($20$\%) was held out for validation. This allowed us to assess the consistency and robustness of the learning process across subsets and to confirm that the training data was informative and of sufficient quality to enable effective model convergence.

Following the cross-validation phase, we applied the fully trained model to a separate held-out test set of $112$ patients. This final evaluation was used to quantify the model’s real-world generalization performance and to ensure that the Acoustic Index remains predictive on unseen clinical data.

All studies included in this dataset were independently reviewed and labeled by board-certified cardiologists. 
Each case was annotated based on structured echocardiographic reports and accompanying clinical information, indicating either the presence of a specific cardiac pathology or the absence of disease in disease-negative individuals. 

\subsection{Dataset Summary}

The dataset used in this study comprises a total of $736$ patients ($10\,053$ DICOM files) 
, including $624$ used for training and $112$ reserved for held-out testing. All patients were over the age of $16$. Among them, $496$ were female and the remaining $240$ were male. Each study contains $2$D/Doppler ultrasound sequences ($30$–$60$ Hz frame rate) across five standardized views: 
\begin{enumerate}
    \item Parasternal long-axis view (PLAX).
    \item Parasternal short-axis mid-papillary view (PSAX-MP).
    \item Parasternal short-axis aortic valve view (PSAX-AV).
    \item Apical 4-chamber view (A4C).
    \item Apical 2-chamber view (A2C).
\end{enumerate}

Table~\ref{tab:age_stats} summarizes key age statistics for both training and test subsets. The age distribution is relatively broad in both sets, with training ages ranging from $16$ to $96$ years (median $68$), and test ages from $16$ to $95$ years (median $72$). The quartile ranges confirm a representative spread across middle-aged and elderly populations.

\begin{table}[!ht]
\centering
\caption{Descriptive age statistics by cohort}
\label{tab:age_stats}
\begin{tabular}{lccccccc}
\hline
\textbf{Cohort} & \textbf{Mean} & \textbf{SD} & \textbf{Min} & \textbf{Max} & \textbf{Median} & \textbf{Q1} & \textbf{Q3} \\
\hline
Training (n = 624) & 65.98 & 17.48 & 18 & 96 & 68 & 54 & 80 \\
Test (n = 112)     & 65.70 & 17.81 & 18 & 95 & 72 & 52 & 79 \\
\hline
\end{tabular}
\end{table}

The dataset encompasses a diverse set of cardiac pathologies commonly encountered in echocardiographic practice. The major disease categories present include Amyloid Cardiomyopathy, Aortic Stenosis, Dilated Cardiomyopathy, Hypertrophic Cardiomyopathy, Ischemic Heart Disease, Non-Ischemic Heart Disease, Valvulopathy.

Data anonymization adhered to the European Union’s General Data Protection Regulation (GDPR) and the European Society of Cardiology’s imaging guidelines, ensuring removal of all patient identifiers while preserving spatiotemporal pixel dynamics.

\subsection{Model Performance on Test Data}

To assess the predictive performance of the Acoustic Index, we conducted a comprehensive evaluation of the trained model on both validation and test data. The following subsections present three complementary analyses. First, we report cross-validation results based on ROC curves and AUC values to quantify the overall data discrimination capability. Next, we analyze the model's sensitivity and specificity across varying thresholds, highlighting the trade-offs involved in clinical deployment. Finally, we visualize the distribution of Acoustic Index scores across patients to explore how well the model separates disease-negative from disease-positive individuals in a continuous risk spectrum.

\subsubsection{ROC Performance}

Receiver Operating Characteristic (ROC) curves are a standard tool for evaluating the discriminative ability of binary classifiers. They plot the true positive rate (sensitivity) against the false positive rate (1 - specificity) at varying decision thresholds. The area under the curve (AUC) summarizes overall performance, with values closer to 1.0 indicating better discrimination between Disease-negative and diseased patients.

\begin{figure}[!ht]
    \centering
    \includegraphics[width=0.9\textwidth]{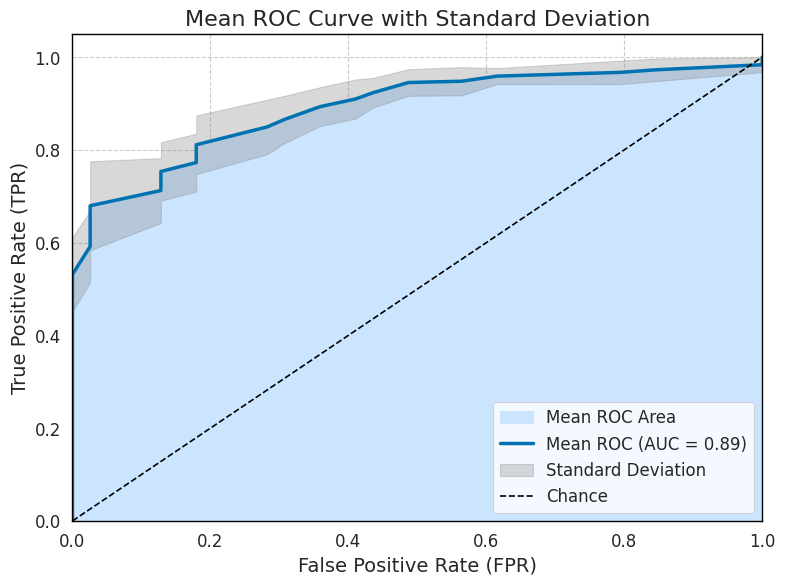}
    \caption{Receiver Operating Characteristic (ROC) curves showing the performance of models trained across five cross-validation folds and evaluated on a common test set. The mean ROC curve and the standard deviation band reflect variability due to different training folds, while evaluation is consistently performed on the same test data.}
    \label{fig:mean_roc}
\end{figure}

The ROC curve in Figure~\ref{fig:mean_roc} shows the average performance across the five cross-validation folds, along with the standard deviation. The model demonstrates consistently high discriminative ability, with area under the curve (AUC) values ranging from 0.88 to 0.91. These results confirm the stability and robustness of the neural network across the entire dataset. The low variability in ROC shape and AUC across folds suggests that the Acoustic Index generalizes well during internal validation and is not overly sensitive to specific data partitions. This consistency supports the reproducibility of the model’s performance under different training conditions.

\subsubsection{Threshold-Based Risk Analysis}

\begin{figure}[!ht]
    \centering
    \includegraphics[width=\textwidth]{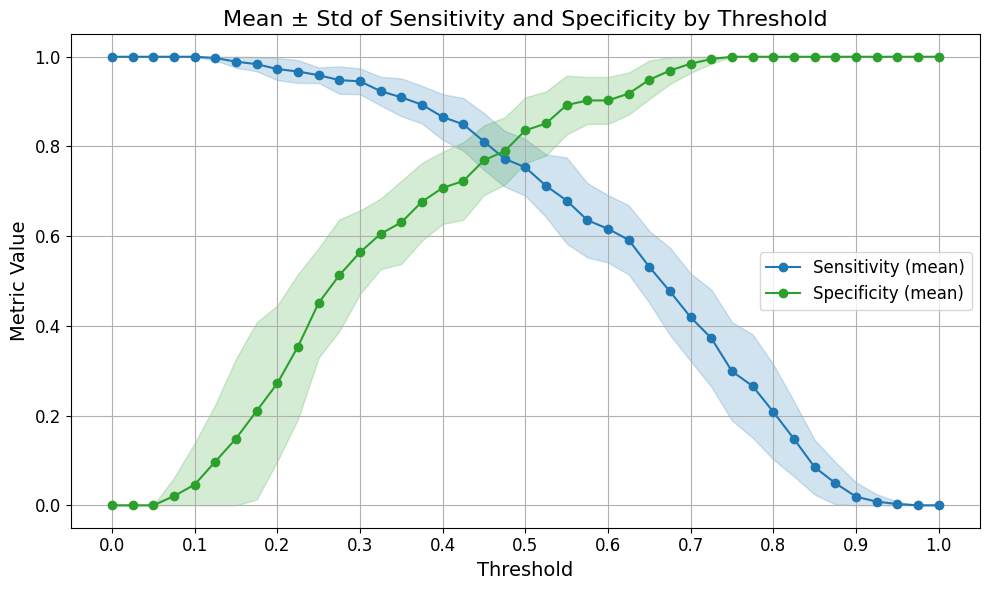}
    \caption{Sensitivity and specificity as functions of the Acoustic Index threshold, computed on a single test set using models trained on five different cross-validation folds. The curves show the mean and standard deviation across folds, reflecting variability in model behavior due to different training data splits.}
    \label{fig:mean_sen_spec}
\end{figure}

Threshold-dependent analysis was performed to assess how the model’s sensitivity (true positive rate) and specificity (true negative rate) vary with the Acoustic Index threshold. Figure~\ref{fig:mean_sen_spec} displays the mean sensitivity and specificity curves across the five cross-validation folds, along with their standard deviation. As expected, increasing the threshold improves specificity while reducing sensitivity, and vice versa, reflecting the inherent trade-off between false positives and false negatives.

This trade-off is clinically relevant, as the choice of threshold determines whether the model favors identifying more diseased patients (high sensitivity) or reducing false positives (high specificity). The intersection point of the sensitivity and specificity curves typically occurs within a threshold range of 0.4 to 0.6, most often below 0.5, with the exception of one fold. This behavior supports the use of the Acoustic Index as a continuous risk score, allowing clinicians to adjust the threshold depending on clinical context, such as prioritizing early detection or minimizing overdiagnosis. 

\subsubsection{Visualizing the Acoustic Index}

\begin{figure}[!ht]
    \centering
    \includegraphics[width=\textwidth]{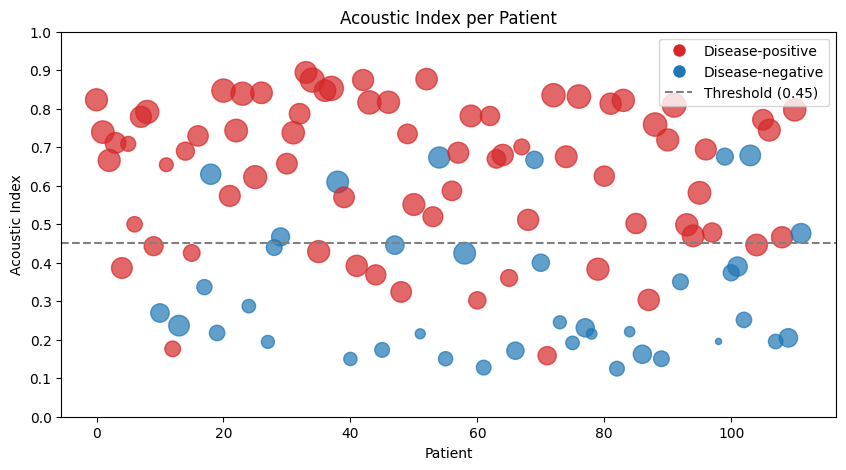}
    \caption{Distribution of Acoustic Index scores for the 112 test patients. Patients are colored by true diagnostic status (red: Disease-positive, blue: Disease-negative), and circle size reflects patient age (circle size increases with age). The dotted line indicates the classification threshold (0.45).}
    \label{fig:vc1_patients}
\end{figure}

The distribution of Acoustic Index scores across the 112 patients in the test dataset is shown in Figure~\ref{fig:vc1_patients}. Each dot corresponds to a patient, color-coded by true diagnostic label (red: Disease-positive, blue: Disease-negative), with circle size proportional to age, with larger circles corresponding to older patients. The model achieves between 77\% and 81\% classification accuracy across different folds, demonstrating the model’s ability to consistently distinguish between patients at high and low risk of cardiac disease.

Age plays a particularly important role: elderly patients tend to exhibit reduced dynamic cardiac capacity, which may lead the model to assign higher Acoustic Index values even in the absence of overt structural abnormalities. This is illustrated in the plot, where Acoustic Index values tend to increase with age (as indicated by circle size), reflecting the real-world observation that older hearts generally present diminished functional and structural performance.

Conversely, younger individuals typically receive lower scores unless they present visible morphological alterations. Although age is included in the model as an input feature, its weight relative to spatiotemporal cardiac patterns is limited, ensuring that the final risk score is not dominated by age alone.

Notably, most false negatives with low acoustic number (Disease-positive patients predicted as Disease-negative) correspond to younger individuals with preserved function, while false positives (Disease-negative patients above the threshold) are often older adults whose normal age-related functional decline may be mistaken for pathology. This reflects both the challenge and subtlety of distinguishing age-related adaptation from early disease, even in human interpretation.

\section{Discussion}

The results presented in this study demonstrate that the proposed Acoustic Index is capable of reliably characterizing cardiac function by integrating both structural and dynamic features extracted from echocardiographic video sequences. Across five cross-validation folds, the model consistently achieved strong performance, with AUC values ranging from 0.88 to 0.91. This level of accuracy is particularly notable given the heterogeneity of the dataset and the simplicity of the final output: a continuous score between 0 and 1.


Compared to conventional metrics such as ejection fraction or Global Longitudinal Strain (GLS), which focus on specific mechanical aspects of the heart, the Acoustic Index offers a more holistic and adaptable measure of cardiac health. Its data-driven formulation allows it to capture complex spatiotemporal patterns without relying on manual annotations or geometric modeling. Moreover, its continuous nature enables flexible threshold selection depending on the clinical application, ranging from population-level screening to prioritizing high-risk patients in real time.

In addition to its flexibility as a continuous score, the Acoustic Index can be further stratified into clinically meaningful regions. For example, a diagnostic “gray zone” may be defined around the classification threshold to capture cases with ambiguous or borderline risk levels. This uncertain region can be determined based on different criteria, such as a fixed interval around the threshold, statistical misclassification limits, or expert-derived ranges, and can be adapted depending on the intended clinical use. In high sensitivity applications like early screening, a wider gray zone may help prioritize follow-up testing, whereas in high specificity scenarios such as surgical decision making, stricter boundaries may be preferred.

The predictions also highlight an important bias: older patients tend to receive higher index values, even when labeled clinically as healthy. This reflects the natural age-related decline in cardiac function and points to the need for careful interpretation of risk scores in elderly populations. At the same time, the model's misclassifications tend to be clinically plausible, such as healthy older adults near the threshold and young sick individuals scoring low, indicating that the learned representations may capture functional realism more than diagnostic labels.

\subsection*{Limitations}

While the results are promising, several limitations must be acknowledged. First, the dataset used in this study, though carefully curated and labeled by cardiologists, is relatively small, which may limit the generalization. External validation on multi-institutional datasets and across different imaging devices is necessary to confirm the robustness of the Acoustic Index.

Second, diagnostic labels were assigned by a single reviewer, which introduces the possibility of inter-observer variability and annotation bias. Future studies should include multiple expert annotations per patient to assess consensus and mitigate labeling noise.

Finally, the current model is designed for global cardiac risk estimation and does not provide differential diagnosis or disease-specific classification. This restricts its interpretability in complex clinical scenarios involving multiple pathologies.
\section{Conclusion \& Future Directions}

This work introduces the Acoustic Index, a novel AI-derived parameter designed to characterize the human heart by integrating both structural (static) and functional (dynamic) information extracted from echocardiographic sequences. The index provides a continuous value between 0 and 1, enabling cardiac risk stratification beyond binary classification. All data used in this study were obtained from clinically validated echocardiographic exams performed by certified cardiologists.

The Acoustic Index is formally defined in Section~\ref{sec:methodology}, where we detail the neural architecture and filtering techniques used to extract spatiotemporal cardiac representations. Despite being trained and validated on a relatively limited dataset of 736 patients, the model demonstrated promising performance, achieving sensitivity and specificity values near or above 80\% in multiple validation folds.

An important feature of the Acoustic Index is its flexibility: the decision threshold can be adjusted depending on the clinical context. For example, thresholds may be lowered to prioritize sensitivity in screening scenarios, or raised in settings where specificity is more critical. This adaptability also enables applications such as automatic patient prioritization, second-opinion support, or intelligent triage in high-throughput settings.

Overall, the Acoustic Index shows strong potential as a clinically useful tool for non-invasive cardiac risk stratification. Rather than replacing clinical decision-making, it is best positioned as an AI-powered assistant, complementing expert judgment by providing a consistent and interpretable measure of cardiovascular function.

\subsection*{Future Work}

The development of the Acoustic Index opens several promising avenues for future research, including its refinement, specialization, and the exploration of novel clinical applications. 

\begin{itemize}
    \item \textbf{Development of specific Acoustic Index applications:} The Acoustic Index can be adapted to address specific clinical tasks beyond global risk stratification. For example, tailored versions of the index could be optimized for early detection, used for integrating the index into targeted screening protocols or decision support tools.

    \item \textbf{Disease-specific refinement:} Extend the model to not only detect global cardiac risk but also to differentiate among specific diseases (e.g., ischemic heart disease, cardiomyopathies).
    
    \item \textbf{Future Risk Prediction} Explore whether the Acoustic Index can serve as a prognostic parameter, identifying patients who are currently asymptomatic or subclinical but may develop cardiac disease in the future. This would involve longitudinal studies to correlate baseline index values with future clinical outcomes and assess its utility in early risk prediction and preventive cardiology.

    \item \textbf{Data expansion and labeling:} Improve model robustness by including additional patient data, covering a wider variety of pathologies and demographic profiles. Labeling the same cases by multiple cardiologists (e.g., 10–20 experts) could help quantify inter-observer variability and reduce bias.
    
    \item \textbf{Multi-level target training:} Future models may benefit from training with a more fine-grained target structure (e.g., scores of 0, 0.5, and 1), allowing the model to explicitly learn from borderline or uncertain cases instead of strict binary labeling.
\end{itemize}

\bibliographystyle{abbrvnat}
\bibliography{main}

\begin{thebibliography}{20}
\providecommand{\natexlab}[1]{#1}
\providecommand{\url}[1]{\texttt{#1}}
\expandafter\ifx\csname urlstyle\endcsname\relax
  \providecommand{\doi}[1]{doi: #1}\else
  \providecommand{\doi}{doi: \begingroup \urlstyle{rm}\Url}\fi

\bibitem[Akkus et~al.(2021)Akkus, Aly, Attia, Lopez-Jimenez, Arruda-Olson, Pellikka, Pislaru, Kane, Friedman, and Oh]{Akkus2021}
Z.~Akkus, Y.~H. Aly, I.~Z. Attia, F.~Lopez-Jimenez, A.~M. Arruda-Olson, P.~A. Pellikka, S.~V. Pislaru, G.~C. Kane, P.~A. Friedman, and J.~K. Oh.
\newblock Artificial intelligence (ai)-empowered echocardiography interpretation: A state-of-the-art review.
\newblock \emph{Journal of Clinical Medicine}, 10\penalty0 (7):\penalty0 1391, 3 2021.
\newblock \doi{10.3390/jcm10071391}.

\bibitem[Al~Kindi et~al.(2021)Al~Kindi, Househ, Alam, and Shah]{AlKindi2021}
D.~Al~Kindi, M.~Househ, T.~Alam, and Z.~Shah.
\newblock Artificial intelligence models for heart chambers segmentation from 2d echocardiographic images: A scoping review.
\newblock In \emph{Informatics and Technology in Clinical Care and Public Health}, volume 289 of \emph{Studies in Health Technology and Informatics}, pages 264--267. IOS Press, 2021.
\newblock \doi{10.3233/SHTI210910}.

\bibitem[Alsharqi et~al.(2018)Alsharqi, Woodward, and Mumith]{Alsharqi2018}
M.~Alsharqi, W.~J. Woodward, and J.~A. Mumith.
\newblock Artificial intelligence and echocardiography.
\newblock \emph{Echo Research and Practice}, 5:\penalty0 R115--R125, 2018.
\newblock \doi{10.1530/ERP-18-0056}.

\bibitem[Arafati et~al.(2020)Arafati, Morisawa, Avendi, Amini, Assadi, Jafarkhani, and Kheradvar]{Arafati2020}
A.~Arafati, D.~Morisawa, M.~R. Avendi, M.~R. Amini, R.~A. Assadi, H.~Jafarkhani, and A.~Kheradvar.
\newblock Generalizable fully automated multi-label segmentation of four-chamber view echocardiograms based on deep convolutional adversarial networks.
\newblock \emph{Journal of The Royal Society Interface}, 17\penalty0 (169):\penalty0 20200267, 2020.
\newblock \doi{10.1098/rsif.2020.0267}.
\newblock URL \url{https://royalsocietypublishing.org/doi/abs/10.1098/rsif.2020.0267}.

\bibitem[Assadi et~al.(2024)Assadi, Alabed, and Li]{Assadi2024}
H.~Assadi, S.~Alabed, and R.~Li.
\newblock Development and validation of ai-derived segmentation of four-chamber cine cardiac magnetic resonance.
\newblock \emph{European Radiology Experimental}, 8:\penalty0 77, 2024.
\newblock \doi{10.1186/s41747-024-00477-7}.

\bibitem[Balaji et~al.(2015)Balaji, Subashini, and Chidambaram]{Balaji2015}
G.~N. Balaji, T.~S. Subashini, and N.~Chidambaram.
\newblock Cardiac view classification using speed up robust features.
\newblock \emph{Indian Journal of Science and Technology}, 8\penalty0 (Supplementary 7):\penalty0 1--5, 2015.
\newblock \doi{10.17485/ijst/2015/v8iS7/62245}.

\bibitem[Brady et~al.(2023)Brady, King, Murphy, and Walsh]{Brady2023}
B.~Brady, G.~King, R.~T. Murphy, and D.~Walsh.
\newblock Myocardial strain: a clinical review.
\newblock \emph{Irish Journal of Medical Science}, 192\penalty0 (4):\penalty0 1649--1656, 8 2023.
\newblock \doi{10.1007/s11845-022-03210-8}.
\newblock Epub 2022 Nov 16.

\bibitem[Gao et~al.(2017)Gao, Li, Loomes, and Wang]{Gao2017}
X.~Gao, W.~Li, M.~Loomes, and L.~Wang.
\newblock A fused deep learning architecture for viewpoint classification of echocardiography.
\newblock \emph{Information Fusion}, 36:\penalty0 103--113, 2017.
\newblock ISSN 1566-2535.
\newblock \doi{https://doi.org/10.1016/j.inffus.2016.11.007}.
\newblock URL \url{https://www.sciencedirect.com/science/article/pii/S1566253516301385}.

\bibitem[Groun et~al.(2022)Groun, Villalba-Orero, Lara-Pezzi, Valero, Garicano-Mena, and {Le Clainche}]{Groun2022}
N.~Groun, M.~Villalba-Orero, E.~Lara-Pezzi, E.~Valero, J.~Garicano-Mena, and S.~{Le Clainche}.
\newblock Higher order dynamic mode decomposition: From fluid dynamics to heart disease analysis.
\newblock \emph{Computers in Biology and Medicine}, 144:\penalty0 105384, 2022.
\newblock ISSN 0010-4825.
\newblock \doi{https://doi.org/10.1016/j.compbiomed.2022.105384}.
\newblock URL \url{https://www.sciencedirect.com/science/article/pii/S0010482522001767}.

\bibitem[Groun et~al.(2023)Groun, Begiashvili, Valero, Garicano-Mena, and {Le Clainche}]{Groun2023}
N.~Groun, B.~Begiashvili, E.~Valero, J.~Garicano-Mena, and S.~{Le Clainche}.
\newblock Higher order dynamic mode decomposition beyond aerospace engineering.
\newblock \emph{Results in Engineering}, 20:\penalty0 101471, 2023.
\newblock ISSN 2590-1230.
\newblock \doi{https://doi.org/10.1016/j.rineng.2023.101471}.
\newblock URL \url{https://www.sciencedirect.com/science/article/pii/S2590123023005984}.

\bibitem[Ouyang et~al.(2020)Ouyang, He, Ghorbani, Yuan, Ebinger, Langlotz, Heidenreich, Harrington, Liang, Ashley, and Zou]{Ouyang2020}
D.~Ouyang, B.~He, A.~Ghorbani, N.~Yuan, J.~Ebinger, C.~P. Langlotz, P.~A. Heidenreich, R.~A. Harrington, D.~H. Liang, E.~A. Ashley, and J.~Y. Zou.
\newblock Video-based ai for beat-to-beat assessment of cardiac function.
\newblock \emph{Nature}, 580:\penalty0 252--256, 3 2020.
\newblock \doi{10.1038/s41586-020-2145-8}.

\bibitem[Park et~al.(2007)Park, Zhou, Simopoulos, Otsuki, and Comaniciu]{Park2007}
J.~H. Park, S.~K. Zhou, C.~Simopoulos, J.~Otsuki, and D.~Comaniciu.
\newblock Automatic cardiac view classification of echocardiogram.
\newblock In \emph{2007 IEEE 11th International Conference on Computer Vision}, pages 1--8, 2007.
\newblock \doi{10.1109/ICCV.2007.4408867}.

\bibitem[Reddy et~al.(2023)Reddy, Lopez, Ouyang, Zou, and He]{Charitha2023}
C.~D. Reddy, L.~Lopez, D.~Ouyang, J.~Y. Zou, and B.~He.
\newblock Video-based deep learning for automated assessment of left ventricular ejection fraction in pediatric patients.
\newblock \emph{Journal of the American Society of Echocardiography}, 36\penalty0 (5):\penalty0 482--489, 2023.
\newblock ISSN 0894-7317.
\newblock \doi{https://doi.org/10.1016/j.echo.2023.01.015}.
\newblock URL \url{https://www.sciencedirect.com/science/article/pii/S0894731723000688}.

\bibitem[Reynaud et~al.(2021)Reynaud, Vlontzos, Hou, Beqiri, Leeson, and Kainz]{Reynaud2021}
H.~Reynaud, A.~Vlontzos, B.~Hou, A.~Beqiri, P.~Leeson, and B.~Kainz.
\newblock Ultrasound video transformers for cardiac ejection fraction estimation.
\newblock In M.~de~Bruijne, P.~C. Cattin, S.~Cotin, N.~Padoy, S.~Speidel, Y.~Zheng, and C.~Essert, editors, \emph{Medical Image Computing and Computer Assisted Intervention -- MICCAI 2021}, pages 495--505, Cham, 2021. Springer International Publishing.

\bibitem[Samad et~al.(2019)Samad, Ulloa, Wehner, Jing, Hartzel, Good, Williams, Haggerty, and Fornwalt]{Manar2019}
M.~D. Samad, A.~Ulloa, G.~J. Wehner, L.~Jing, D.~Hartzel, C.~W. Good, B.~A. Williams, C.~M. Haggerty, and B.~K. Fornwalt.
\newblock Predicting survival from large echocardiography and electronic health record datasets: Optimization with machine learning.
\newblock \emph{JACC: Cardiovascular Imaging}, 12\penalty0 (4):\penalty0 681--689, 2019.
\newblock ISSN 1936-878X.
\newblock \doi{https://doi.org/10.1016/j.jcmg.2018.04.026}.
\newblock URL \url{https://www.sciencedirect.com/science/article/pii/S1936878X18303851}.

\bibitem[Sanchez-Martinez et~al.(2018)Sanchez-Martinez, Duchateau, Erdei, Kunszt, Aakhus, Degiovanni, Marino, Carluccio, Piella, Fraser, and Bijnens]{SanchezMartinez2018}
S.~Sanchez-Martinez, N.~Duchateau, T.~Erdei, G.~Kunszt, S.~Aakhus, A.~Degiovanni, P.~Marino, E.~Carluccio, G.~Piella, A.~G. Fraser, and B.~H. Bijnens.
\newblock Machine learning analysis of left ventricular function to characterize heart failure with preserved ejection fraction.
\newblock \emph{Circulation: Cardiovascular Imaging}, 11\penalty0 (4):\penalty0 e007138, 2018.
\newblock \doi{10.1161/CIRCIMAGING.117.007138}.
\newblock URL \url{https://www.ahajournals.org/doi/abs/10.1161/CIRCIMAGING.117.007138}.

\bibitem[Strait and Lakatta(2012)]{Strait2012}
J.~B. Strait and E.~G. Lakatta.
\newblock Aging-associated cardiovascular changes and their relationship to heart failure.
\newblock \emph{Heart Failure Clinics}, 8\penalty0 (1):\penalty0 143--164, 1 2012.
\newblock \doi{10.1016/j.hfc.2011.08.011}.

\bibitem[Williams et~al.(2015)Williams, Kevrekidis, and Rowley]{Williams2015}
M.~O. Williams, I.~G. Kevrekidis, and C.~W. Rowley.
\newblock A data--driven approximation of the koopman operator: Extending dynamic mode decomposition.
\newblock \emph{Journal of Nonlinear Science}, 25\penalty0 (6):\penalty0 1307--1346, 2015.
\newblock ISSN 1432-1467.
\newblock \doi{10.1007/s00332-015-9258-5}.
\newblock URL \url{https://doi.org/10.1007/s00332-015-9258-5}.

\bibitem[Zhang et~al.(2019)Zhang, Yu, and Li]{Zhang2019}
W.~Zhang, Y.-C. Yu, and J.-S. Li.
\newblock Dynamics reconstruction and classification via koopman features.
\newblock \emph{Data Mining and Knowledge Discovery}, 33\penalty0 (6):\penalty0 1710--1735, 11 2019.
\newblock \doi{10.1007/s10618-019-00639-x}.
\newblock Epub 2019 Jun 24.

\bibitem[Zhao et~al.(2024)Zhao, Wang, Wong, and Wang]{Zhao2024}
D.~Zhao, Y.~Wang, N.~D. Wong, and J.~Wang.
\newblock Impact of aging on cardiovascular diseases: From chronological observation to biological insights: Jacc family series.
\newblock \emph{JACC Asia}, 4\penalty0 (5):\penalty0 345--358, 4 2024.
\newblock \doi{10.1016/j.jacasi.2024.02.002}.

\end{thebibliography}

\end{document}